\documentclass[letterpaper, 10 pt, conference]{ieeeconf}
\usepackage{times}
\usepackage[pdftex]{graphicx}
\usepackage{subfigure}
\usepackage{amsmath,amssymb,amsopn,amstext,amsfonts}
\usepackage{cancel}
\usepackage[space]{cite}
\usepackage{pdfsync}
\usepackage{balance}
\usepackage{color}
\usepackage{mathtools}
\usepackage{algpseudocode}
\usepackage[ruled,vlined,linesnumbered]{algorithm2e}

\algnewcommand\algorithmicforeach{\textbf{for each}}
\algdef{S}[FOR]{ForEach}[1]{\algorithmicforeach\ #1\ \algorithmicdo}

\usepackage{bm}

\usepackage{diagbox}
\usepackage{epstopdf}
\usepackage{pifont}
\usepackage{fixltx2e}
\usepackage{amsmath}
\usepackage{multirow}
\usepackage{url}
\usepackage{verbatim}
\usepackage{footmisc}
\usepackage[linkcolor=black,citecolor=black,urlcolor=black,colorlinks=true]{hyperref}
\usepackage{subfigure}
\usepackage{color}
\usepackage[skip=3pt, font={small}]{caption} 
\usepackage{tabularx}

\graphicspath{{figures/}}
\DeclareGraphicsExtensions{.png,.jpg,.eps,.pdf}
\IEEEoverridecommandlockouts
\title{\LARGE \bf
Loam\_livox: A fast, robust, high-precision LiDAR odometry and mapping package for LiDARs of small FoV
}

\author{Jiarong Lin and Fu Zhang 
\thanks{J. Lin and F. Zhang are with the Department of Mechanical Engineering, Hong Kong University, Hong Kong SAR., China. {\tt\small $\{$jiarong.lin,  fuzhang$\}$@hku.hk}
}
}%

\begin{document}
\maketitle

\newcommand{\note}[1]{\textcolor{red}{\emph{\bf#1}}}
\newcommand\footnoteref[1]{\protected@xdef\@thefnmark{\ref{#1}}\@footnotemark}
\newlength{\bibitemsep}\setlength{\bibitemsep}{.0238\baselineskip}
\newlength{\bibparskip}\setlength{\bibparskip}{0pt}
\let\oldthebibliography\thebibliography
\renewcommand\thebibliography[1]{%
	\oldthebibliography{#1}%
	\setlength{\parskip}{\bibitemsep}%
	\setlength{\itemsep}{\bibparskip}%
}
\begin{abstract}
LiDAR odometry and mapping (LOAM) has been playing an important role in autonomous vehicles, due to its ability to simultaneously localize the robot's pose and build high-precision, high-resolution maps of the surrounding environment. This enables autonomous navigation and safe path planning of autonomous vehicles. In this paper, we present a robust, real-time LOAM algorithm for LiDARs with small FoV and irregular samplings. By taking effort on both front-end and back-end, we address several fundamental challenges arising from such LiDARs, and achieve better performance in both precision and efficiency compared to existing baselines. To share our findings and to make contributions to the community, we open source our codes on Github\footnote{\url{https://github.com/hku-mars/loam_livox}\label{loam_livox}}.

\end{abstract}

\section{Introduction}\label{sect_intro}

With the ability to provide long range, highly accurate 3D measurements of the surrounding environment, light detection and ranging (LiDARs) is becoming an essential sensor in many robotic applications, such as autonomous driving vehicles \cite{levinson2011towards}, drones \cite{bry2012state, gao2019flying}, surveying, and mapping \cite{nuchter20076d,schwarz2010lidar}. To enable massive use in these areas, recent developments in LiDAR technologies have been focusing on lowering the device cost while increasing its reliability \cite{ieeespect}. In this trend, one class of LiDARs that gains increasingly interests and developments are solid state LiDARs, which come with various implementations, such as micro-electro-mechanical-system (MEMS) scanning, optical phase array (OPA), Risley prism, etc. Being massively produced \footnote{\label{mid40} \url{ https://www.livoxtech.com/mid-40-and-mid-100}}, these high performance and extremely low-cost LiDARs hold the potential to promote or radically change the robotics industry.
	
Despite their superiority in cost, reliability, and possibly performance against the conventional mechanical spinning LiDARs, such as Velodyne Puck \footnote{\url{https://velodynelidar.com/vlp-16.html}}, solid state LiDARs have many new features that bring significant challenges to the LiDAR navigation and mapping. These features are (to explain these features, we take the Livox MID40 LiDAR \footref{mid40} as an example due to its wide availability): 

\begin{figure}[t]
	\centering
	{\includegraphics[width=1.0\columnwidth]{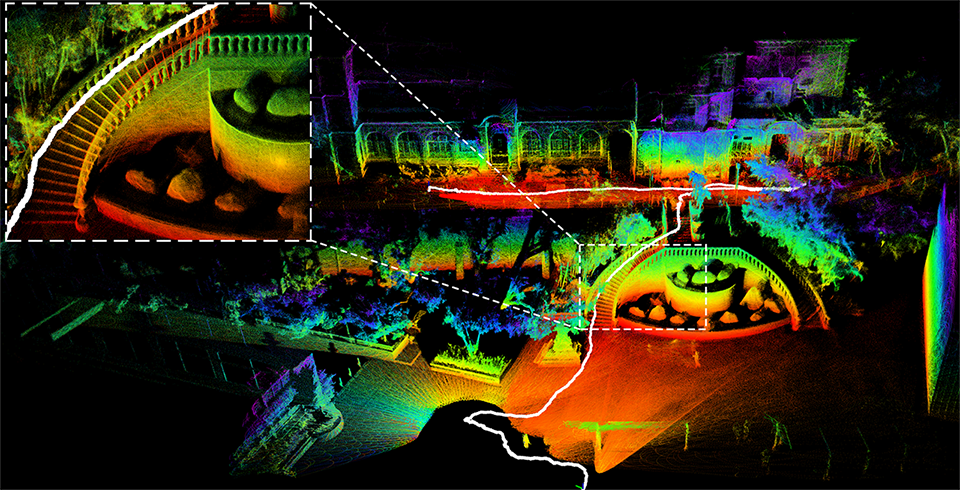}}\\
	\caption{The 3D map of the Chong Yuet Ming Cultural Center in the University of Hong Kong (HKU).}
	\label{fig_hku_zym}
	\vspace{0.05cm}
	\includegraphics[width=1.0\columnwidth]{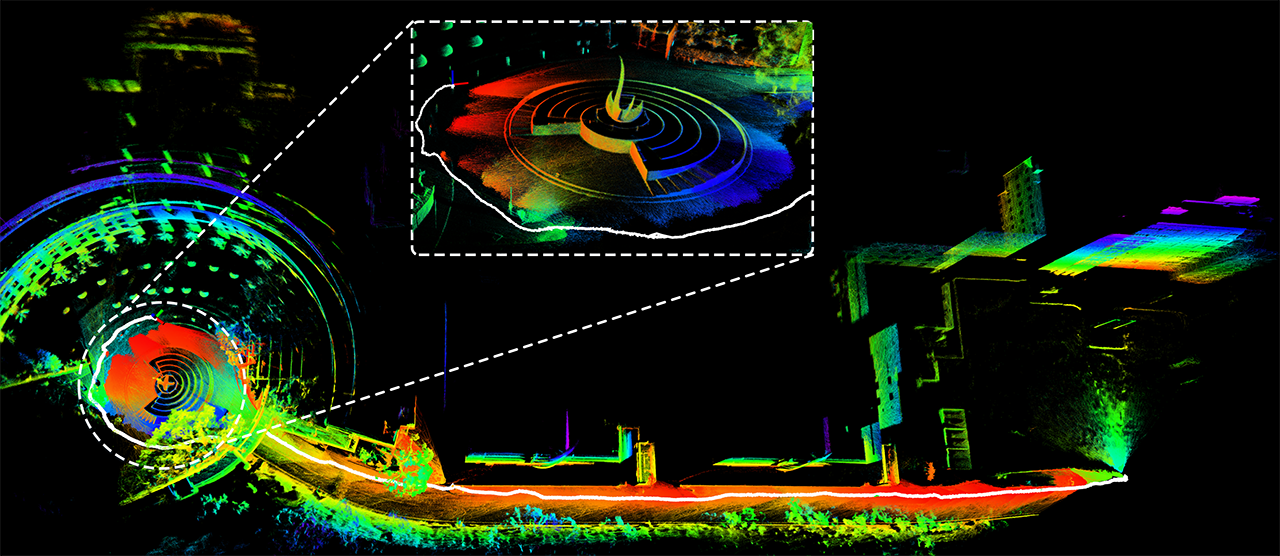}\\
	\vspace{0.05cm}
	\includegraphics[width=1.0\columnwidth]{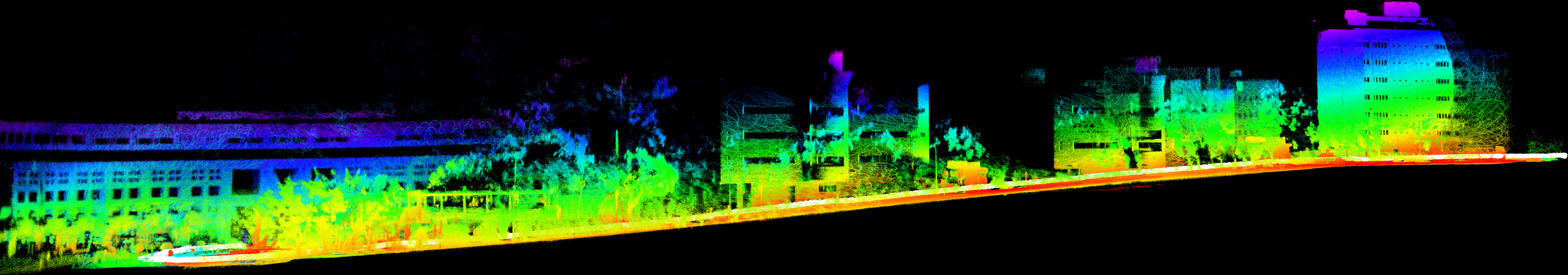}
	\caption{The large scale mapping of the Hong Kong University of Science and Technology (HKUST) main campus, the upper and lower images are the bird-view and side-view, respectively. In above images, the white path is the trajectory of the LiDAR, points are colored by their heights.}
	\label{fig_hkust}
	\vspace{-0.8cm}
\end{figure}

\begin{figure}[t]
	\centering
	\vspace{-0.0cm}
	\includegraphics[width=1.0 \columnwidth]{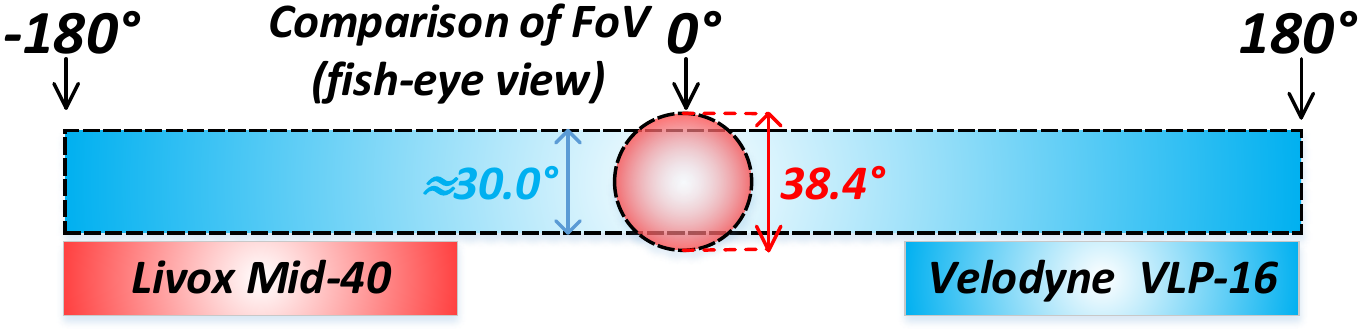}
	\caption{FoV of Livox Mid-40 and Velodyne PUCK (VLP-16).}
	\label{fig_fov_comp}
	\vspace{-1.0cm}
\end{figure}

\textit{{Small FoV:}}  solid state LiDARs usually have very small field of view (FoV). For examples, Livox MID40 has a front facing, conical shaped FoV spanning 38.4 degrees. Other solid state LiDARs such MEMS LiDARs also suffer from similar small FoV problem due to the large size of the MEMS mirror preventing large steering angles. Comparing to conventional spinning LiDAR (see Fig. \ref{fig_fov_comp}), the reduced FoV will lead to very fewer features in a frame, making the subsequent feature matching prone to degenerate and easily disturbed by moving objects. Although a larger FoV can be obtained by combining multiple LiDARs, it considerably increases the sensor cost and weight. 

\begin{figure}[t]
	{\includegraphics[width= 1.0  \columnwidth]{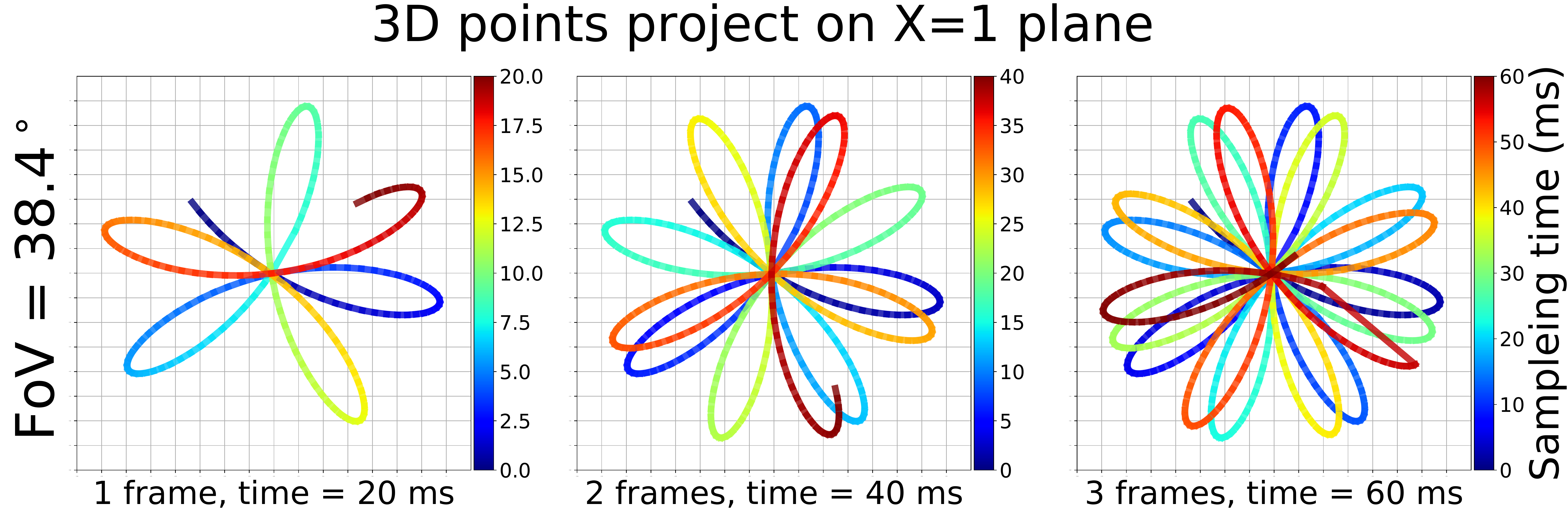}}
	\caption{The scanning trajectory of 3D points projected on the plane of $ 1m $ distance in front, where the color encodes the sampling time.}
	\label{fig_scan_pattern}
	\vspace{-1.7cm}
\end{figure}

\textit{{Irregular scanning pattern:}}  existing spinning LiDARs have multiple laser-receiver pairs stacking in a vertical row. Rotating all pairs as a whole leads to a collection of parallel rings. This regular scanning greatly simplifies the feature extraction. For example, a corner is easily computed by differentiating the depth of points on the line.  In constrat, the scanning pattern of solid state LiDARs is quite irregular. For example, the Livox MID40 has a rosette-like scanning pattern (see Fig.~\ref{fig_scan_pattern}) where two neighboring scanning petals are separated far apart.

\textit{{Non-repetitive scanning:}} to maximize the coverage ratio even when the LiDAR is static, non-repetitive scanning is usually adopted \cite{mid40pc} where the scanning trajectory never repeats itself (see Fig. \ref{fig_scan_pattern}).

\textit{{Motion blur:}} due to the continuous scanning of a single laser head, the 3D points measured in one frame are really sampled at different times as the LiDAR is continuously moving. The in-frame motion will distort the point clouds and cause motion blur.

To address the problems mentioned above, we develop a software package named ``Loam\_Livox'', which addresses many key issues including feature extraction and selection in a very limited FoV, robust outliers rejection, moving objects filtering and motion distortion compensation. Without other sensors such as IMU, GPS, and cameras, our algorithm calculates the LiDAR poses in real time (i.e. odometry) by registering its point cloud to a specified range of local map. Some of the results we obtained are shown in Fig.~\ref{fig_hku_zym} and Fig.~\ref{fig_hkust}, where we can tell the precision of the algorithm from the level of details of the stairs and railing (Fig.~\ref{fig_hku_zym}), as well as versatility for large-scale mapping (Fig.~\ref{fig_hkust}).

\section{Related work}


State estimation and map-building are the fundamental prerequisites for intelligent robots. In the past recent years, we have seen great efforts being made in the field of simultaneous localization and mapping (SLAM), including both vision-based and laser-based approaches. In this paper, we mainly focus on the problem of laser-based SLAM.

Besl \textit{et al} \cite{besl1992method} first proposed the Iterative {Closest} Points (ICP) method for scan matching, which builds the basic operation for odometry. Building on this, Mendes \textit{et al} \cite{mendes2016icp} proposed a pose-graph SLAM to correct the drift in sequential scan matching, and demonstrated its effectiveness in a high definition LiDARs, Velodyne HDL 64.

While the ICP algorithm performs well for 3D scans with dense points, its effectiveness considerably degrades when the points in a scan are sparse where the two scans do not scan the same location on an object. To solve this problem, Pulli \textit{et al }\cite{pulli1999multiview} proposed a “point-to-plane” error metric. This metric is used together with the “point-to-point” metric in \cite{besl1992method} and called the generalized ICP in \cite{segal2009generalized}. Zhang \textit{et al} \cite{zhang2014loam} and Shan \textit{et al}\cite{shan2018lego} also used the “point-to-edge” metric in the  context of LiDAR odometry and mapping.

Besides the geometry features mentioned above, 3D keypoints based method \cite{rusu2008learning, rusu2009fast, rusu2010fast} have also been proposed. These methods required less computation resources, by extracting keypoints from dense point cloud with detector like Point Feature Histograms (PFH) \cite{rusu2008learning, rusu2009fast}, Viewpoint Feature Histograms (VFH) \cite{rusu2010fast}, etc. Considering the point cloud characteristic of our scenarios and the demands of real-time performance, we use point-to-edge and point-to-plane feature in our work,  inspired by the work of \cite{zhang2014loam, shan2018lego}.

To eliminate the effects of motion blur caused by LiDAR movement, authors in \cite{zhang2014loam} , \cite{ hong2010vicp} and \cite{droeschel2018efficient} compensate the movements in front-end processing by linear interpolating the LiDAR pose. More recently, Gentil {\it et al} \cite{gentil2019in2laama} formulates an optimization problem in the back-end processing to compensate the LiDAR movement. Compared to the previous work, the back-end processing method achieves better performance but cannot run in real-time.

While most of the previous work were based on spinning LiDARs, in this work, we focus on the odometry and mapping with solid-state LiDARs of small FOVs. Our contributions are: (1) we develop a complete LOAM algorithm for LiDARs with small FOVs. The algorithms is  carefully engineered and made open source to benefit the community; (2) we increase the accuracy and robustness of the LOAM algorithm by considering the ow-level physical properties of LiDAR sensors in the front-end processing; (3) we propose a simple yet effective motion compensation method, the piecewise processing, and parallelize its implementation. Experiments show that the piecewise processing outperforms linear interpolation in terms of accuracy and running efficiency. 

\section{Points selection and feature extraction}

\begin{figure*}[htp]
	\centering
	\includegraphics[width=2.1\columnwidth]{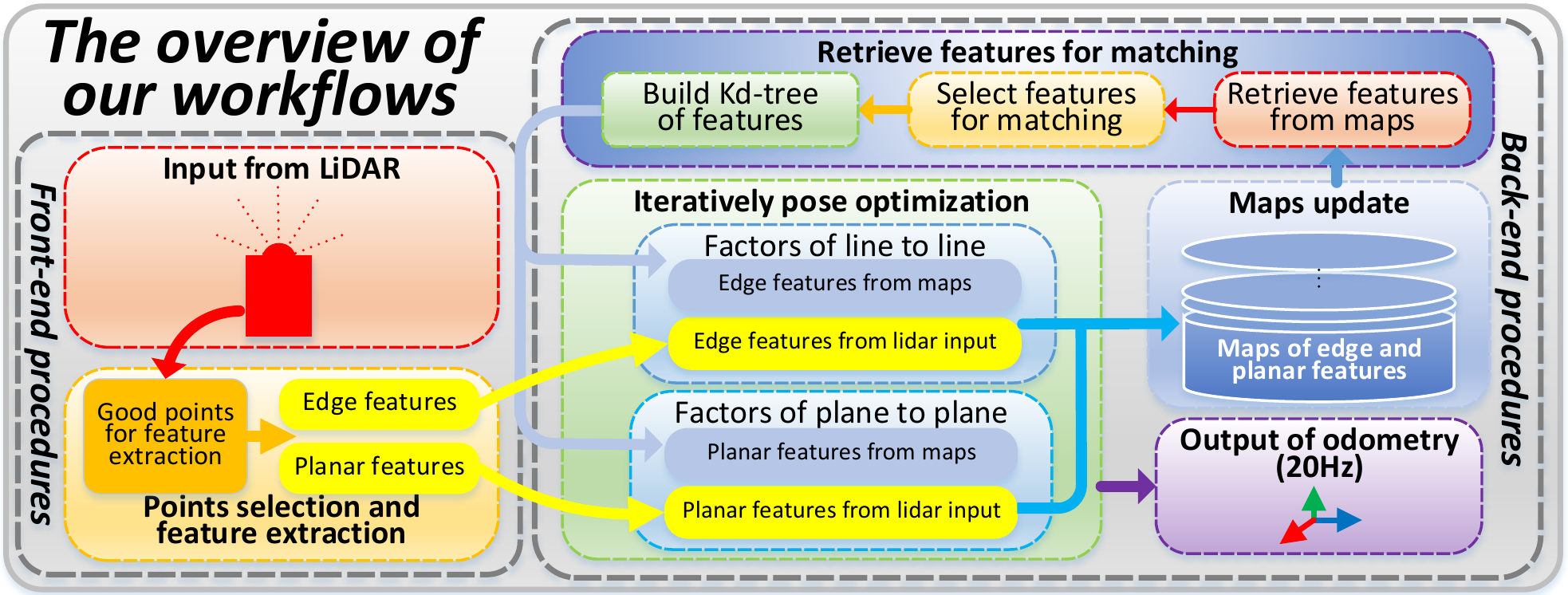}
	\caption{The overview of our workflows. Each new frame is matched directly with the global map to {provide} the odometry output. The matching result is in turn used to register the frame to the global map, leading to the same rate (i.e., 20 Hz) of odometry output and map update. In our implementation, only the feature points (i.e., edge points and plane points) are saved in memory and all the raw points are saved in hard disk for possible offline processing (e.g., offline global optimization).}
	\label{fig_workflow}
	\vspace{-0.5cm}
\end{figure*}

The overview or our system is shown as Fig.~\ref{fig_workflow}, whose front-end processing comprises of the point selection and feature extraction. Considering the measuring mechanism of a LiDAR sensor its low-level physical properties (e.g., laser spot size, signal noise ratio), we perform a point level selection to extract the ``good points''. 

\subsection{Points selection}
\begin{figure}
	\centering
	\includegraphics[width = 1.0\columnwidth]{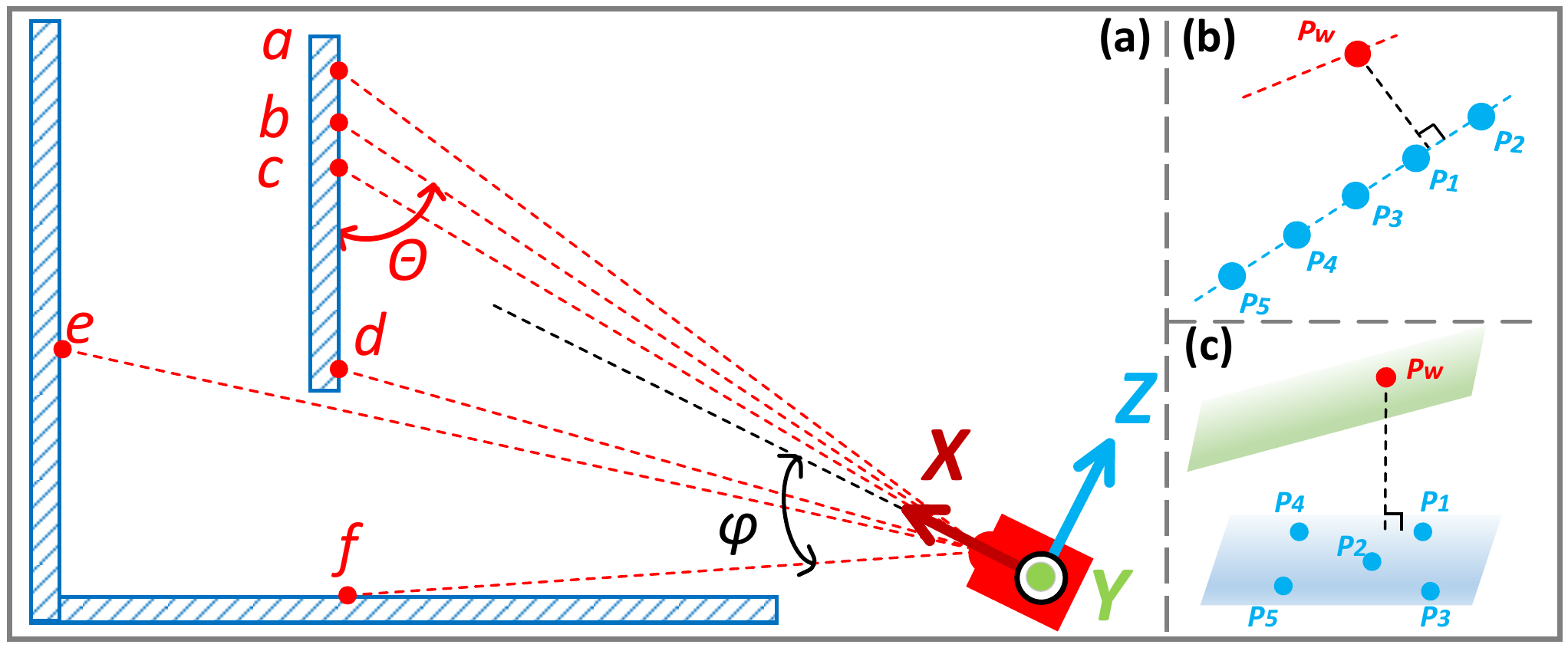}\\
	\caption{(a) Illustration of incident angle $\theta$, deflection angle $\phi$; (b) residual of edge-to-edge;  (c) residual of plane-to-plane.}
	\label{fig_point_selection}
	\vspace{-0.6cm}
\end{figure}

We compute the following features of each 3D point $ \mathbf{P} = \left[x,y,z\right] $, where the $X-Y-Z$ axis correspond to the Front-Left-Up (FLU) of a LiDAR (see Fig.~\ref{fig_point_selection} (a)). 

	Depth $D$ is the distance of the measured point to the LiDAR sensor.
	\begin{equation}
	\label{eq_depth}
		D(\mathbf{P}) = \sqrt{x^2+y^2+z^2}
	\end{equation}


	The laser deflection angle $\phi$ is the angle between $X$ axis and laser ray	
	\begin{equation}
	\label{eq_phi}
		\phi(\mathbf{P}) =  \tan^{-1}\left( \sqrt{{(y^2+z^2)} / {x^2}} \right)
	\end{equation}
	
	The intensity $I$ is
	\begin{equation}
	\label{eq_intensity}
		I(\mathbf{P}) = R / D(\mathbf{P})^2
	\end{equation}
	where $R$ is the object reflectivity, measured by the LiDAR sensor (some LiDARs, e.g., Velodyne Puck, returns the intensity instead of reflectivity. In this case, the intensity is directly available). Small intensity $ I(\mathbf{P}) $ means the point is either far from the LiDAR sensor (large  $ D(\mathbf{P}) $) or the object reflectivity $ R $ is low.

	The incident angle $\theta$ is the angle between the laser lay and the local plane around the measured point  (Fig.~\ref{fig_point_selection} (a)).
	\begin{equation}
	\theta(\mathbf{P}_b) = \cos^{-1} \left( \dfrac{ (\mathbf{P}_a- \mathbf{P}_c)\cdot  \mathbf{P}_b   }{ \left| \mathbf{P}_a- \mathbf{P}_c \right| \left| \mathbf{P}_b \right|  }\right)	
	\end{equation}

To increase the localization and mapping accuracy, we remove any of the following points: 
 \begin{itemize}
	\item[1.] Points nearing to the fringe of the FoV. (e.g., $\phi(\mathbf{P}) \geq 17^\circ $ for Livox MID40). In such area, scanning trajectory has large curvatures, leading to the feature extraction in Section III.B less reliable. 
	
	\item[2.] Points with too large or too small intensity (e.g. $I(\mathbf{P}) \leq 7\times 10^{-3} \text{,or } 	I(\mathbf{P}) \geq 1\times 10^{-1} $ for MID40 ). This is because intensity directly indicates the strength of the received laser signal. Too large intensity (signal) usually leads to saturation or distortion in the receiving circuitry and decreases the ranging accuracy. On the other hand, too small intensity (signal) usually leads to lower signal noise ratio, which also deteriorate the ranging accuracy.
	
	\item[3.] Points with incident angles near to $\pi$ or $0$ (e.g. $\theta(\mathbf{P}) \leq 5^\circ \text{,or } \theta(\mathbf{P}) \geq 175^\circ$ for MID40), like point $ \mathbf{P}_f $ in Fig.~\ref{fig_point_selection} (a). This is because the laser spot caused by the nonzero divergence angle of the laser beam will be considerably elongated. As a result, the measured range is the average of the area covered by the large spot instead of a specific point.
	\item[4.] Points hidden behind an objects (e.g., $\mathbf{P}_e$ in Fig.~\ref{fig_point_selection} (a)), which will cause a false edge feature otherwise. A point $\mathbf{P}_e$ is a hidden point if: $$| \mathbf{P}_e  - \mathbf{P}_d |  \geq 0.1 | \mathbf{P}_e | \text{, and} |\mathbf{P}_e| > |\mathbf{P}_d|$$ where $\mathbf{P}_d$ is the nearest measurement point in scanning order.
\end{itemize}
\subsection{Feature extraction}\label{sect_feature_extraction}

After points selection, we perform feature extraction to extract features from the ``good  points''. We extract plane and edge features by computing the local smoothness of the point candidate as in \cite{zhang2014loam}.  Furthermore, to mitigate the matching degeneration due to the small number of features caused by the limited FoV and the point selection, we employ the LiDAR reflectivity as the 4-th dimensional measurement. If the reflectivity of a 3D point is considerably different its neighborhood points, we treat it as a point of edge feature (edge in the reflectivity due to materials change, in contrast to the edge in geometry due to shape change). Such points are beneficial in some of the degeneration cases like facing a wall with closed doors and windows.

\section{Iterative pose optimization}

Due to the non-repetitive scanning mentioned in Section. \ref{sect_intro}, the extracted feature cannot be constantly tracked  like in \cite{zhang2014loam, shan2018lego, gentil2019in2laama}. A simple example is that, even when the LiDAR is static, the scanned trajectory (and feature points) are different from the previous frame. In our work, we use an iterative pose optimization procedure to calculate the LiDAR pose. With the proper implementation detailed later, we achieve real-time odometry and mapping, both at 20Hz.

\subsection{Residual of edge-to-edge}\label{sect_e2e}
Denote $\boldsymbol{\mathcal{E}}_k$ and $\boldsymbol{\mathcal{E}}_m$ the set of all edge features (see Section.~\ref{sect_feature_extraction}) in the current frame and in the map, respectively. For each point in  $\boldsymbol{\mathcal{E}}_k$, we find $5$ nearest points from $\boldsymbol{\mathcal{E}}_m$ (see Fig.~\ref{fig_point_selection} (b)). To boost the searching speed, we build a \textit{KD-tree} of  $\boldsymbol{\mathcal{E}}_m $ (see Fig.~\ref{fig_workflow}). Moreover, the KD-tree is built by another parallel thread once the last registered frame/sub-frame is received (see Fig. \ref{fig_parallel_diagram}). This makes the KD-tree immediately available when the new frame is received. 

Let $\mathbf{P}_l$ be a point in $\boldsymbol{\mathcal{E}}_k$ of the current frame ($ k $-th frame). Noticing that the point  $\mathbf{P}_l$  in $\boldsymbol{\mathcal{E}}_k$ is in the local LiDAR frame while points of $\boldsymbol{\mathcal{E}}_m$ are registered in the global map, to find the nearest points of $\mathbf{P}_l$ in $\boldsymbol{\mathcal{E}}_m$, we need to project it to the global map by the following transformation. 
\begin{equation}\label{eq_lidar_to_world}
\mathbf{P}_w = \mathbf{R}_{k} \mathbf{P}_{l} +  \mathbf{T}_{k}
\end{equation}
where $(\mathbf{R}_{k} , \mathbf{T}_{k})$ is the LiDAR pose when the last point in current frame is sampled, and needs to be determined by the pose optimization. Here we use the LiDAR pose at the last point in a frame to represent the pose of the whole frame, and all points in that frame are projected to the global map using this pose. Also notice that the last point in the current frame is essentially the first point in the next frame. 

Let $ \mathbf{P}_i $ denote the $i$-th nearest points of $\mathbf{P}_w$ of $\boldsymbol{\mathcal{E}}_m$. 
To make sure that $ \mathbf{P}_i $ is indeed on a line, we compute the mean $\boldsymbol{\mu} $ and covariance matrix $\mathbf{\Sigma}$ formed by the $ m $ nearest points of $ \mathbf{P}_w $. We set $m$ to $ 5 $ in our work. If the biggest of eigenvalue of $\mathbf{\Sigma}$ is three times larger than the second biggest eigenvalue, we assure that the nearest points of $\mathbf{P}_w$ form a line on which $\mathbf{P}_w$ should lie. The residual is then computed as (Fig. \ref{fig_point_selection} (b)).

\begin{equation}\label{eq_res_e2e}
\mathbf{r}_{e2e} = \dfrac { \left|  \left(\mathbf{P}_w - \mathbf{P}_5\right)\times \left(\mathbf{P}_w - \mathbf{P}_1\right) \right| }{ \left| \mathbf{P}_5 - \mathbf{P}_1  \right| }
\end{equation}
	

\subsection{Residual of plane-to-plane}\label{sect_p2p}

Similar to the edge feature points, for a point in the planar feature set  $\boldsymbol{\mathcal{P}}_k$ of current frame, we find 5 nearest points in the planar feature set $\boldsymbol{\mathcal{P}}_m$ of the map (see Fig.~\ref{fig_point_selection}(c)). We also assure these 5 nearest points are indeed within the same plane by computing their covariance matrix $\Sigma$. If the smallest eigenvalue of $\mathbf{\Sigma}$ is three times less than the second smallest eigenvalue, we compute the distance of the plane point in the current frame to the plane formed by the 5 points in the same plane, as follows, and add this residual to pose optimization.
\begin{equation}\label{eq_res_p2p}
\mathbf{r}_{p2p} = \dfrac{ \left( \mathbf{P}_w - \mathbf{P}_1 \right)^T  \left( \left(\mathbf{P}_3 - \mathbf{P}_5 \right)\times \left(\mathbf{P}_3 - \mathbf{P}_1 \right)    \right) }{ \left| \left(\mathbf{P}_3 - \mathbf{P}_5 \right)\times \left(\mathbf{P}_3 - \mathbf{P}_1 \right) \right| }
\end{equation}

\subsection{In-frame motion compensation}
As mentioned previously, the 3D points are sampled at different time of different poses (i.e., motor blur) as the LiDAR motion is continuously undergoing. To eliminate the effect of motion blur, we propose two methods as follows:

\subsubsection{Piecewise processing}
\begin{figure}[t] 
	\centering
	{\includegraphics[width=1.0\columnwidth]{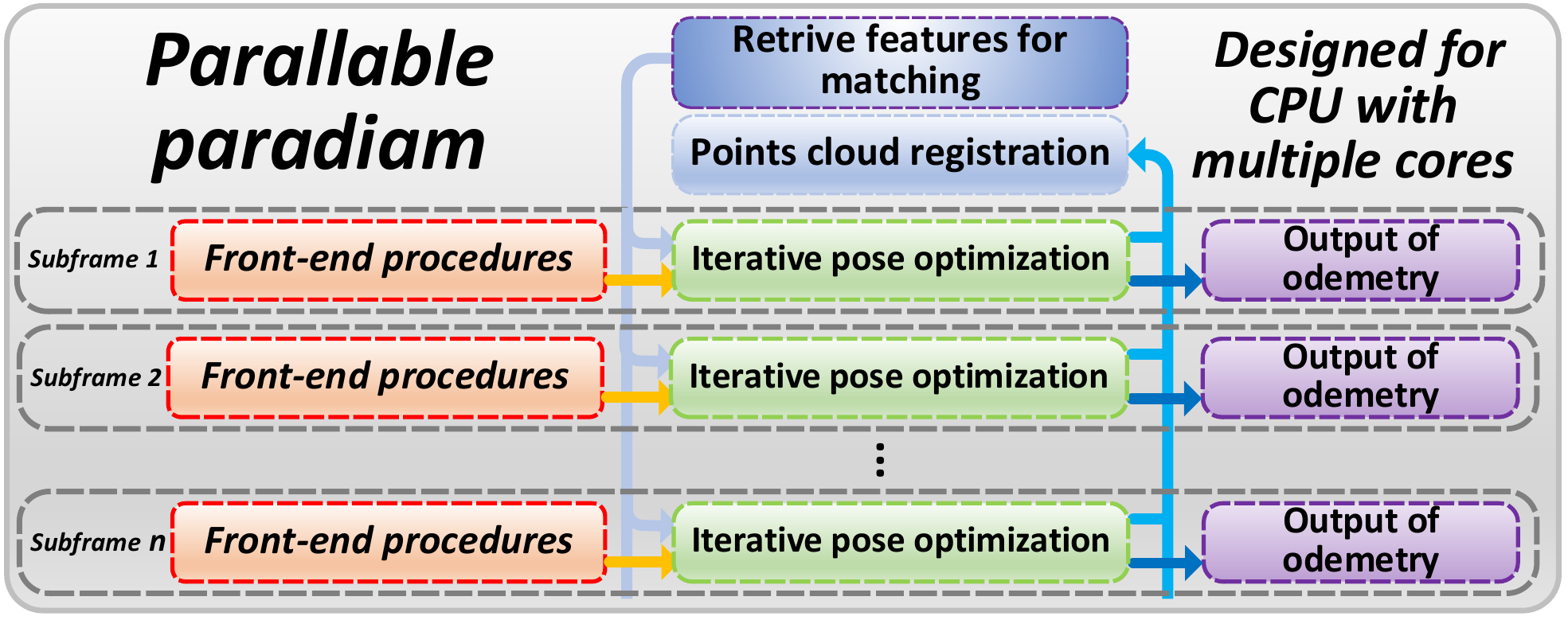}}\\
	\caption{Our parallel paradigm for CPU with multiple cores. Each sub-frame is matched with the global map independently on a dedicated thread. The matched sub-frame is then registered to the global map and become a part of the map. Another dedicated thread receives the new registered sub-frame and build a KD three of the updated map to be used in the next frame. }
	\label{fig_parallel_diagram}
	\vspace{-1.5cm}
\end{figure}

A simple yet effective way to eliminate the effect of motion blur is piecewise processing. We divide an incoming frame into three sequential sub-frames. Then match these three sub-frames to the same map accumulated so far independently. During the scan matching of each sub-frame, all its points are projected to the global map using the LiDAR pose at the end point of that sub-frame. By doing so, the time {interval} of each sub-frame is 1/3 of the original frame. Although this method seems very simple, it works surprisingly well as shown in the results. Additionally, this piecewise processing has the benefits of utilizing the multi-core structure in modern CPUs by parallelizing the matching of each sub-frame (see Fig. \ref{fig_parallel_diagram}). 

\begin{figure*}[t]
	\includegraphics[width=1.95\columnwidth]{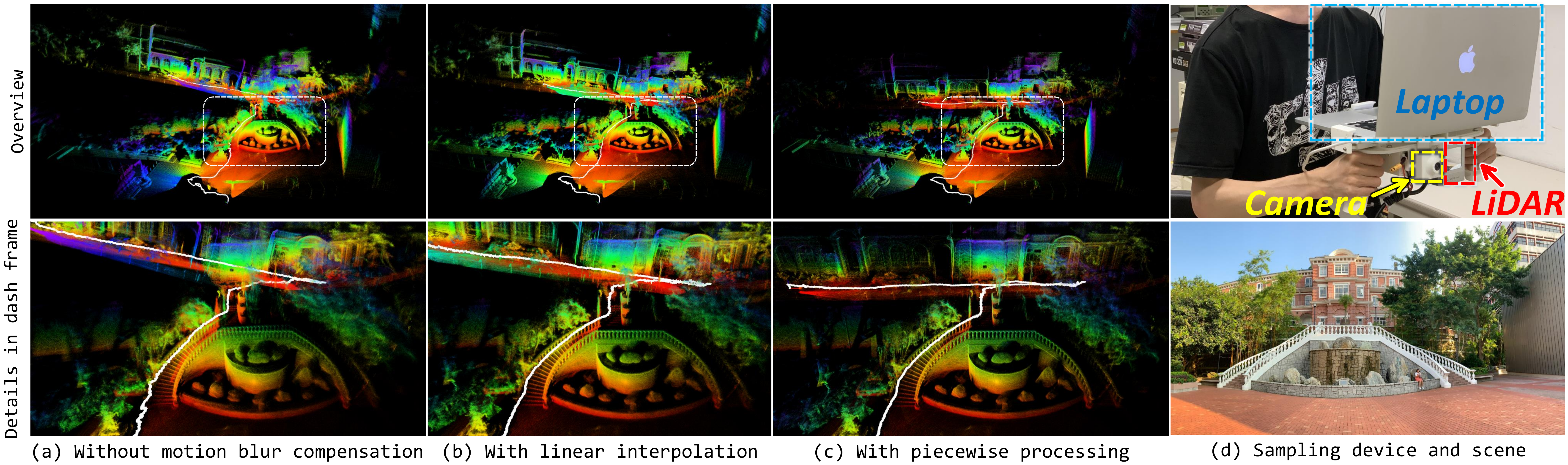}
	\caption{The comparison of different motion compensation methods. The first column shows the results without any motion compensation, the second is with linear interpolation, and the third column is piecewise processing. The upper picture in the fourth column is our hand-held device for data collection, the lower picture is the RGB image of the mapped area.}
	\label{fig_motion_blur_eliminate}
	\vspace{-0.2cm}
\end{figure*}

\subsubsection{Linear interpolation}
Another commonly used motion compensation method is linear interpolation, as in \cite{zhang2014loam, gentil2019in2laama}. Denote $(\mathbf{R}_k , \mathbf{T}_k)$ the LiDAR pose at the last point of the current frame and $(\mathbf{R}_{k-1} , \mathbf{T}_{k-1})$ the previous frame, $(\mathbf{R}_{k-1}^{k}, \mathbf{T}_{k-1}^{k})$ the relative rotation and translation between the previous and the current frame, then:
$$
	\mathbf{R}_{k} =\mathbf{R}_{k-1}\mathbf{R}_{k-1}^{k},~~\mathbf{T}_{k} =\mathbf{R}_{k-1}\mathbf{T}_{k-1}^{k} + \mathbf{T}_{k-1}\\
$$

Assume $ t_{k-1} $ is the sampling time of the last point in the previous frame. For any point sampled at time $ t $ in the current frame, we have $ t  \in [t_{k-1}, t_{k}] $. Compute $s = (t-t_{k-1})/(t_{k} - t_{k-1})$, then, the linearly interpolated pose at time $ t $ is: 
$$
	\mathbf{R}_{k-1}^{t} = e^ {\hat{\boldsymbol{\omega}} \theta s}, ~~\mathbf{T}_{k-1}^{t} = s\mathbf{T}_{k-1}^{k} \\
$$
where $\theta$ is the magnitude and $\boldsymbol{\omega} $ is the unit vector of of the rotation axis of $\mathbf{R}_{k-1}^{k} $, respectively. $ \widehat{\boldsymbol{\omega}} $ is the skew symmetric matrix of $\boldsymbol{\omega} $. From the Rodrigue's formula \cite{murray2017mathematical}, we have:
$$
	\mathbf{R}_{k-1}^{t} = \mathbf{I} + \widehat{\boldsymbol{\omega}}\sin (s\theta) + \hat{\boldsymbol{\omega}}^2(1-\cos (s\theta))
$$
which implies that only $ \sin(s\theta) $ and $ \cos(s\theta) $ needs to be computed for each point of the current frame, while the rests remain constant. This saves some computations. With $\mathbf{R}_{k-1}^{t} $, the LiDAR pose at the current time is: 
\begin{equation}
\label{eq_pose_interpolate}
\mathbf{R}_t = \mathbf{R}_{k-1} \mathbf{R}_{k-1}^t, ~~\mathbf{T}_t = \mathbf{R}_{k-1} \mathbf{T}_{k-1}^{t} + \mathbf{T}_{k-1}
\end{equation}

Then we can project the point at time $ t $ to the global map by the interpolated pose, as follows: 
\begin{equation}
\label{eq_motion}
	\mathbf{P}_w(t) = \mathbf{R}_t \mathbf{P}_l +  \mathbf{T}_t
\end{equation}

\begin{algorithm}[t] 
	\caption{Iterative LiDAR pose optimization}
	\label{alg_it_pose}
	\renewcommand{\thealgocf}{}
	\renewcommand{\theAlgoLine}{}  
	\SetKwInOut{Input}{Input}
	\SetKwInOut{Output}{Output}
	\SetKwInOut{Begin}{Begin}
	\SetKwInOut{Start}{Start}
	\Input{The edge set $\boldsymbol{\mathcal{E}}_{k}$ and plane set $\boldsymbol{\mathcal{P}}_{k}$ from the current (sub-) frame; The edge set $\boldsymbol{\mathcal{E}}_m$ and plane set $\boldsymbol{\mathcal{P}}_m$ from maps; The LiDAR pose of the previous frame $(\mathbf{R}_{k-1} , \mathbf{T}_{k-1}) $.}
	\Output{The pose  of the current frame $( \mathbf{R}_{k} , \mathbf{T}_{k}) $. }
	\Start
	{	$\mathbf{R}_{k} \leftarrow \mathbf{R}_{k-1}$,
		$\mathbf{T}_{k} \leftarrow \mathbf{T}_{k-1}$}
	\For{Iterative pose optimization is not converged}
	{
		\For{ $\mathbf{P}_l\in \boldsymbol{\mathcal{E}}_{k}$ }
		{
			Compute $\mathbf{P}_w$ via (\ref{eq_lidar_to_world}) (or (\ref{eq_motion})).\\
			Find 5 nearest points $\left\{ \mathbf{P}_{1\sim5} \right\}$ of $\mathbf{P}_w$ in $\boldsymbol{\mathcal{E}}_m$.
			
			\If{ $\left\{ \mathbf{P}_{1-5} \right\}$ are indeed in a line}
			{
				Add edge-to-edge residual $\mathbf{r}_{e2e}$ via (\ref{eq_res_e2e}).
			}
		}
		\For{ $\mathbf{P}_l\in \boldsymbol{\mathcal{P}}_{k}$ }
		{
			Compute $\mathbf{P}_w$ via (\ref{eq_lidar_to_world}) (or (\ref{eq_motion})).\\
			Find 5 nearest points $\left\{ \mathbf{P}_{1\sim5} \right\}$ of $\mathbf{P}_w$ in $\boldsymbol{\mathcal{P}}_m$.
			
			\If{ $\left\{ \mathbf{P}_{1\sim5} \right\}$ are indeed a plane}
			{
				Add plane-to-plane residual $\mathbf{r}_{p2p}$ via (\ref{eq_res_p2p}).
			}
		}
		Perform pose optimization with 2 iterations.\\
		Recompute $\mathbf{r}_{e2e}$ and $\mathbf{r}_{p2p}$, then remove $20\%$ of the biggest residual.
		
		\For{a maximal number of iterations}
		{
			\If{the nonlinear optimization converges}
			{Break;}	
		}
	}
	\vspace{-0.1cm}
\end{algorithm}

\subsection{Outliers rejection, dynamic objects filtering}

To avoid moving object in real environments {bringing} down the accuracy of scan matching, we perform a dynamic objects filtering as follows: in each iteration of the iterative pose optimization, we refind the nearest {neighbors} of each feature point and add the edge-to-edge residual (\ref{eq_res_e2e}) and plane-to-plane residual (\ref{eq_res_p2p}) to the objective function, we first perform pose optimization with a small number of iterations (e.g., 2 used in our experiments). Using the optimization results, we compute the two residuals in (\ref{eq_res_e2e}) and (\ref{eq_res_p2p}), and remove the first $20\%$ largest residuals. With the outliers removed, a full pose optimization is finally performed. The complete iterative pose optimization algorithm is summarized in Algorithm.~\ref{alg_it_pose}.

\section{Results}

\subsection{Evaluation of mapping}
The comparisons of the two motion compensation methods of are shown in  Fig.~\ref{fig_motion_blur_eliminate}, where we can see that, without any motion compensation (Fig.~\ref{fig_motion_blur_eliminate} (a)), the mapping is very {blurry} in local areas (e.g., stairs, railing) and distorted in larger scale (e.g., the building is curved). In contrast, with the motion compensation, both of the linear interpolation and piecewise processing effectively eliminate the motion blur, and the stair steps and railing are distinguishable one from another. However, the linear interpolation has a considerable long-term drift, as seen by the curved building in the upper figure of Fig.~\ref{fig_motion_blur_eliminate} (b). This is because the data are collected by hand-held devices and the movement could be quite jerky and cannot be accurately captured by simple linear interpolation. 

\subsection{Evaluation of odometry}

\begin{figure}[t]
	\centering
	{\includegraphics[width=1.0\columnwidth]{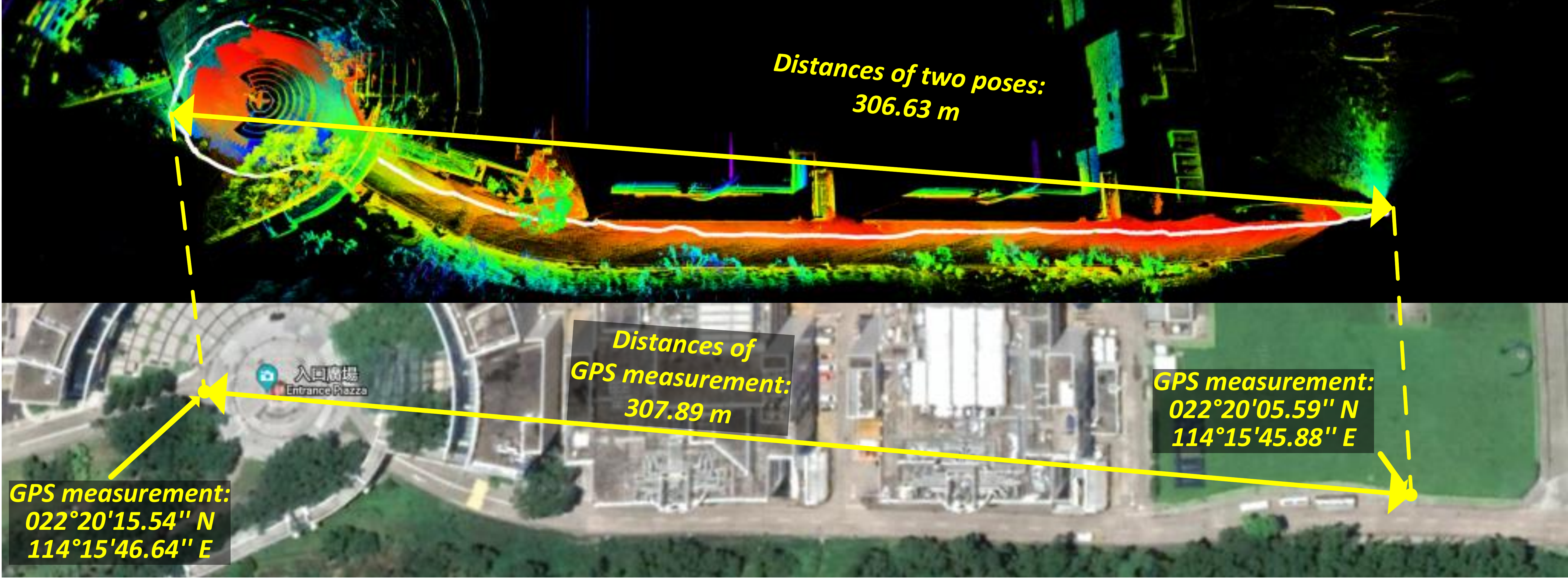}}
	\vspace{0.3cm}
	{\includegraphics[width=1.0\columnwidth]{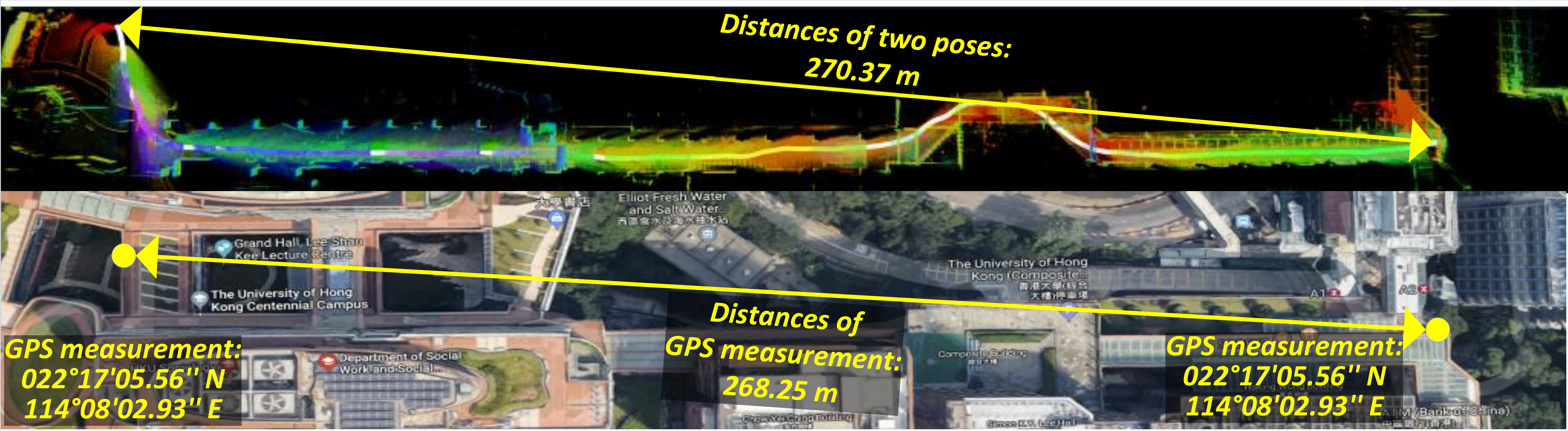}}
	\caption{The localization accuracy on two datasets: outdoor (upper) and indoor (lower). In each dataset, we compare our odometry results with Google maps and compute the traveled distance.}
	\label{fig_eval_precision}
	\vspace{-0.1cm}
\end{figure}

We evaluate the localization of our algorithm by comparing with the measurement of GPS, shown in Fig.~\ref{fig_eval_precision}. We compute the distance of two positions of our odometry and then compare it with the measurement of GPS. The results on two datasets are $0.41 \%$ and $0.65 \%$, respectively, implying that the localization is of high accuracy. 

\begin{figure}[t]
	\centering
	{\includegraphics[width=0.95\columnwidth]{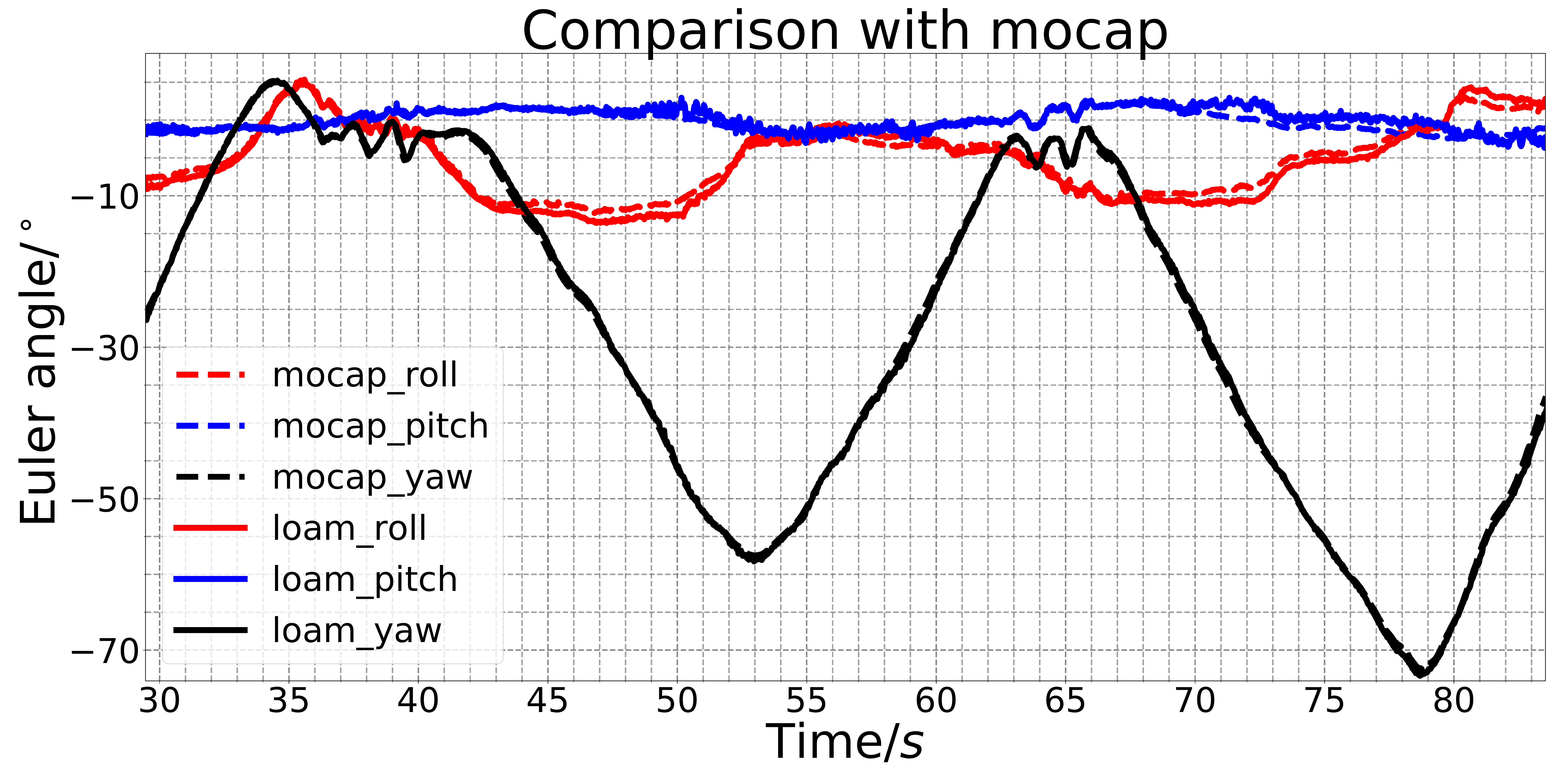}}\\
	\caption{The comparison between our results and motion capture (mocap) system, the dashed line is measured by mocap, and the solid line is the odometry output from our algorithm.}
	\label{fig_comp_mocap}
	\vspace{-1.0 cm}
\end{figure}

Furthermore, we evaluate the accuracy of rotation by comparing our result with {the} motion capture system (mocap) shown in Fig.~\ref{fig_comp_mocap}). The results show that the trajectories of our odometry and mocap are very close and the average error of Euler angles in all three directions is as low as $1.1^\circ$.

\subsection{Evaluation of running performance}
We evaluate the time consumption of our algorithm and the current baseline \footnote{\url{https://github.com/HKUST-Aerial-Robotics/A-LOAM}\label{aloam}} (both algorithm eliminates the motion blur with piecewise processing) on two platforms, the desktop PC (with \textit{i7-9700K}) and onboard-computer (DJI manifold2 \footnote{\url{https://www.dji.com/cn/manifold-2}} with \textit{i7-8550U}). As shown in Table.~\ref{tab_time_profile}, benefiting from the parallelization among sub-frame registration, as well as between feature matching and KD-tree building, our algorithms run \textit{2$\sim$3} times faster than the baseline.



\begin{table}[t]
{
\setlength{\extrarowheight}{.1em}
\setlength\extrarowheight{0.01pt}
\begin{tabular}[h]{|c|c|c|c|c|}
	\hline
 		 & \scriptsize{Desktop PC}& \scriptsize{Desktop PC} &\scriptsize{Onboard PC}&\scriptsize{Onboard PC}\\
		 & \scriptsize{@4.0$\sim$4.8 Ghz} & parallel & \scriptsize{@3.0$\sim$3.5 Ghz} & parallel\\
	\hline
	Ours& 35.68 ms & 17.24 ms & 54.60 ms & 32.54 ms \\
	\hline
	Baseline& 109.00 ms & NaN & 125.13 ms & NaN \\
	\hline
\end{tabular}
\caption{The time consumption for each frame of our algorithm and the baseline\footref{aloam}, where ``Desktop PC parallel'' and ``Onboard PC parallel'' use 3 threads for point cloud registration.}
\label{tab_time_profile}
\vspace{-0.5cm}
}
\end{table}

\subsection{Others}
Due to the space limit, we strongly recommend the reader to review our code \footnote{\url{https://github.com/hku-mars/loam_livox}} and more results contained there \footnote{\url{https://github.com/ziv-lin/loam_livox_paper_res}}.

\section{Conclusion and discussion}
This paper presented an odometry and mapping algorithm for LiDARs with small FOVs. The algorithm inherits the basic structure and techniques (e.g., feature extraction, mating, motion compensation by linear interpolation) of standard LOAM algorithm, but with several key new contributions, such as point selection, iterative pose optimization, and implementation parallelization.  The developed algorithm has its odometry and mapping both running in real time (i.e., 20 Hz). While achieving a high level of accuracy in mapping and localization, the sequential scan matching is inherently drifting. Reducing this drift by using techniques like loop closure and sliding window optimization will be further researched in the future.

\bibliography{icra2020jiarong}

\end{document}